\documentclass{elsarticle-1}

\usepackage[utf8]{inputenc} 
\usepackage[T1]{fontenc}    
\usepackage{hyperref}       
\usepackage{url}            
\usepackage{booktabs}       
\usepackage{amsfonts}       
\usepackage{nicefrac}       
\usepackage{microtype}      
\usepackage{xcolor}         
\usepackage{algorithm}		
\usepackage{algcompatible}
\usepackage{subcaption}     
\usepackage{graphicx}

\usepackage{natbib}
\usepackage{mathtools}
\usepackage{amsmath,amssymb,amsbsy,amsfonts,latexsym,
               amsopn,amstext,amsxtra,euscript,amscd,commath,mathtools, braket, physics, hyperref}

\newcommand{\x}{\text{x}}

\usepackage{multirow}

\newtheorem{theorem}{Theorem}[section]
\newtheorem{definition}{Definition}[section] 
\def\bproof{\textbf{Proof. }}
\def\eproof{\hfill$\Box$}

\def\cB{{\mathcal B}}

\def\cD{{\mathcal D}}

\def\cF{{\mathcal F}}

\def\cH{{\mathcal H}}

\def\cS{{\mathcal S}}

\def\cX{{\mathcal X}}
\def\cY{{\mathcal Y}}

\def\bb {{\bf b}}

\def\xx  {{\bf x}}
\def\yy{{\bf y}}

\def\00{{\bf 0}}
\def\11{{\bf 1}}
\def\+{\oplus}

\begin{document}

\title{On Learning with  LAD}
\author[1]{C.~A.~Jothishwaran}
\ead{jc\_a@ee.iitr.ac.in}
\affiliation[1]{organization={Department of Electronics and Communication Engineering},
addressline={Indian Institute of Technology Roorkee},
city={Roorkee},
postcode={247667},
country={India}}

\author[2]{Biplav Srivastava}
\ead{biplav.s@sc.edu} 
\affiliation[2]{organization={Department of Computer Science and Engineering, AI Institute College of Engineering and Computing},
addressline={University of South Carolina},
city={Columbia SC},
postcode={29208},
country={USA}}

\author[3,4]{Jitin Singla} 
\ead{jsingla@bt.iitr.ac.in}
\affiliation[3]{organization={Department of Biosciences and Bioengineering},
addressline={Indian Institute of Technology Roorkee},
city={Roorkee},
postcode={247667},
country={India}}

\author[4]{Sugata Gangopadhyay}
\ead{sugata.gangopadhyay@cs.iitr.ac.in}
\affiliation[4]{organization={Department of Computer Science and Engineering},
addressline={Indian Institute of Technology Roorkee},
city={Roorkee},
postcode={247667},
country={India}}

\begin{abstract}
The logical analysis of data, LAD, is a technique that yields two-class classifiers based on 
Boolean functions having disjunctive normal form (DNF) representation. 
 Although LAD algorithms employ optimization techniques, 
the resulting binary classifiers 
or binary rules do not lead to overfitting.  We propose a theoretical 
justification for the absence of overfitting by estimating the Vapnik-Chervonenkis dimension 
(VC dimension) for LAD models where hypothesis sets consist of 
DNFs with small number of cubic monomials. We illustrate and confirm our 
observations empirically.  
\end{abstract}

\maketitle


\section{Introduction}
Suppose we have a collection of observations for a particular phenomenon in the form of
data points and the information about its occurrence at each data point. We refer to such a 
data set as the {\it training set}. Data points are (feature) vectors whose coordinates 
are values of variables called {\em features}. The information on the occurrence or 
non-occurrence of the phenomenon under consideration can be recorded by labeling each 
data point as a ``false''  point or a ``true'' point, alternatively, 
by $0$ or $1$, respectively. 
Peter~L.~Hammer~\cite{ham86} proposed using partially defined Boolean functions to 
explore the cause-effect relationship of a data point's membership in the set 
of ``true'' points or ``false'' points. Crama et al.~\cite{cra88} developed 
this theory and named it the {\it Logical Analysis of Data}, or LAD for short. 
Another noteworthy survey article is by Alexe et al.~\cite{ale07}, where the authors 
discuss LAD in detail and focus on using LAD for biomedical data analysis. 

Here, we consider LAD in the Probably Approximately Correct (PAC) learning model 
framework. We denote the hypothesis set by $\cH$. We restrict $\cH$ to the set 
of Disjunctive Normal Forms (DNFs) involving a small number of cubic 
terms and estimate the Vapnik-Chervonenkis (VC) dimension for the hypothesis set. 
Recently, 
Chauhan et al.~\cite{ChauhanMGG22} compared LAD with DNN and CNN for analyzing 
intrusion detection data sets. It was observed that LAD with low-degree terms 
(cubic and degree four) offer classifiers that outperform DNN or CNN classifiers. 
In this article, we theoretically 
explain why we can expect to learn from data using LAD is possible by solely 
checking the accuracy of the proposed Boolean classifiers within the training set.

\section{Partially defined Boolean functions and logical analysis of data}
\label{pdBf}

Let $\mathbb{Z}$ be the ring of integers, and $\mathbb{Z}^+$ be the set of positive 
integers. Consider the set $\cB = \{0, 1\}$. 
For any $u, v \in \cB$, not necessarily distinct, we define 
disjunction, conjunction, and negation as
$u \vee v = u + v -uv$, $u \wedge v = uv$, and $\bar{u} = 1-u$,  
respectively, where the operations on the right-hand side are over $\mathbb{Z}$.  
It is customary to write $uv$ instead of $u \wedge v$. 
The set $\cB=\{0, 1\}$ along with these operations is a Boolean algebra. 
For $n \in \mathbb{Z}^+$, let $[n]=\{1, \ldots, n\} \subset \mathbb{Z}^+$. 
The cartesian product of $n$ copies 
of $\cB$ is $\cB^n = \{\xx = (x_1, \ldots, x_n) : x_i \in \cB, i \in [n]\}$. 
The set $\cB^n$ is a Boolean algebra where disjunction, conjunction, and negation 
are induced from those defined over $\cB$ as: 
$\xx \vee \yy = (x_1 \vee y_1, \ldots, x_n \vee y_n)$, 
        $\xx \wedge \yy = (x_1 \wedge y_1, \ldots, x_n \wedge y_n)$, and 
        $\bar{\xx} =  (\bar{x}_1, \ldots, \bar{x}_n)$, 
for all $\xx, \yy \in \cB^n$. 

Let the set of all functions from a set $\cX$ to a set $\cY$ be denoted by 
$\cF^{\cX, \cY}$. In this paper, $\cX = \cB^n$ and $\cY = \cB$. 
A function $f \in \cF^{\cB^n, \cB}$ is said to be an $n$-variable 
Boolean function. 
The support or the set of true points of $f$  
is $T(f) = \{\xx \in \cB^n : f(\xx) = 1  \}$, 
and the set of false points  is $F(f) = \{\xx \in \cB^n : f(\xx) = 0  \}$. 
An $n$-variable Boolean function can be completely defined by the ordered pair 
of sets $(T(f), F(f))$.  Clearly, $T(f) \cup F(f) = \cB^n$  and $T(f) \cap F(f) = \emptyset$. 
Hammer~\cite{ham86} proposed the notion of partially defined Boolean functions as 
follows. 
\begin{definition}
Let $T, F \subseteq \cB^n$ such that 
$T\cap F = \emptyset$. Then $(T, F)$ is said to be a partially defined 
Boolean function, or pdBf, in $n$ variables. 
\end{definition}
For a pdBf $(T, F)$, it is understood that $T \cup F \neq \cB^n$, otherwise the
pdBf $(T, F)$ is a Boolean function.
For studying Boolean functions and their various applications, we refer to 
\cite{CramaH}. 

This paper considers two-class classification problems with
feature vectors in $\cB^n$. For a positive integer $N$, consider a random sample
of $\cS =\{\xx^{(1)}, \ldots, \xx^{(N)}\}  \subseteq \cB^n$ of size $N$. Let the label 
corresponding to the  $\xx^{(i)}$ be denoted by $y^{(i)} \in \cB$ for all $i\in [N]$. 
The vectors belonging to $\cS$, each augmented with its binary label, form the training set
$\cD = \{(\xx^{(i)}, y^{(i)}) : i \in [N] \}$.  The sets
\begin{equation*}
        T_{\cD} = \{ \xx^{(i)} : y^{(i)} = 1, i \in [N]\}, \mbox{ and }
        F_{\cD} = \{ \xx^{(i)} : y^{(i)} = 0, i \in [N]\}
\end{equation*}
are said to be the sets of positive and negative examples, respectively. 
The pair of subsets $(T_{\cD}, F_{\cD})$ is a partially defined Boolean 
function over $\cB^n$.  
\begin{definition}
A Boolean function $f: \cB^n \rightarrow \cB$ is an {\it extension} of 
a pdBf $(T, F)$, if $T \subseteq T(f)$ and $F \subseteq F(f)$.    
\end{definition}

LAD uses the pdBf $(T_{\cD}, F_{\cD})$
corresponding to a training set $\cD$ and proposes its extension 
as an approximation of the target function. Researchers have demonstrated 
that such extensions, when carefully constructed using particular 
conjunctive rules, provide excellent approximations
of target functions. 
Boros et al.~\cite[page 34, line 7]{DBLP:journals/anor/BorosCHIKM11} 
call them classifiers based on the ``most justifiable'' rules and 
further state that these ``rules do not seem to lead to overfitting, even though
it (the process of finding them) involves an element of optimization.'' 
In this paper, we prove this observation within the framework of the PAC 
learning model. Before proceeding further, we introduce some definitions 
and notations to describe our results. 

A Boolean variable is a variable that can take values from the set $\cB$. 
Let $x$ be a Boolean variable. We associate 
a Boolean variable, $\bar{x}$, with $x$ such that for all $x \in \cB$, 
$x \bar{x} = 0$ and $x \vee \bar{x}  = 1$. The symbol $x^\alpha$ is defined by
$$
x^\alpha  = \begin{cases}
    x &\rm{ if } \;\alpha = 1 \\
    \bar{x} &\rm{ if } \;\alpha = 0. 
\end{cases}
$$
The symbol $x^\alpha$ is said to be a literal. 

A LAD algorithm outputs a collection of prime patterns that maximally cover the 
true points of the pdBf $(T, F)$ obtained from the training set $\cD$. 
For the technical details, we refer to \cite{ham86,cra88,ale07,CramaH} and other 
related research results. 
In this paper, we do not focus on developing efficient algorithms to obtain 
theories and testing for how accurately they approximate a target function. 
Instead, we aim to establish the conditions that make learning by Boolean 
rules feasible. In other words, we would like to understand 
why we do not usually see overfitting even if the LAD algorithms are designed 
to maximally fit a theory with the training set data. We propose to do this analysis
by using the PAC learning model. 

\section{The PAC learning model}
\label{PACmodel}

Valiant~\cite{Valiant84, Valiant-stoc-84} proposed the theory of 
Probably Approximately Correct (PAC) in 1984. For an introduction 
to the concept of the VC dimension, we refer to 
Abu-Mostafa et al.~\cite{mostafa}.
Let us denote the set of all possible feature vectors and labels 
by $\cX = \cB^n$ and $\cY = \cB$, respectively.
We assume that for each phenomenon there is a {\it target function} $f \in \cF^{\cX, \cY}$ 
that correctly labels all the vectors in $\cX$. We consider training sets with binary 
features and labels of the form $\cD = \{(\xx^{(i)}, y^{(i)}): i \in [N] \}$ where 
each $\xx^{(i)} \in \cB^n$ and $y^{(i)} \in \cB$ are data points and binary labels, 
respectively. By definition, the target function $f \in \cF^{\cX, \cY}$ satisfies 
$f(\xx^{(i)}) = y^{(i)}$, for all $i \in [N]$. 
Let $\cH \subset \cF^{\cX, \cY}$ be a set of functions called 
the hypothesis set. The PAC learning involves approximating the target function 
$f \in \cF^{\cX, \cY}$ by a function $h \in \cH \subset \cF^{\cX, \cY}$ such 
that it has the lowest average error for points inside and outside the training 
set $\cD$. The hypothesis set ought to be carefully chosen and fixed before the 
execution of a learning algorithm over a training set. 
\begin{definition}
The in-sample error is the fraction of data points in $\cD$ where the target function 
$f$ and $h \in \cH$ disagree. That is, 
\begin{equation}
\label{in-error}
    E_{\rm{in}}(h) = \frac{1}{N} \sum_{i \in [N]} \#\{\xx^{(i)} : h(\xx^{(i)}) \neq f(\xx^{(i)})\}. 
\end{equation}
\end{definition}
It is realistic to assume that the input space $\cX$ has a probability distribution
$\mu$ defined on it. For an input $\xx$ chosen from this space satisfying the probability 
distribution $\mu$, we write $\xx \sim \mu$. 
The out-of-sample error is the probability that $h(\xx) \neq f(\xx)$ when $\xx \sim \mu$. 
\begin{definition}
The out-of-sample error is 
\begin{equation}
\label{out-error}
    E_{\rm{out}}(h) = \Pr_{\xx \sim \mu}[h(\xx) \neq f(\xx)]. 
\end{equation}
\end{definition}
Learning is feasible if the learning algorithm can produce a function $g \in \cH$ such that 
the in-sample error is close enough to the out-of-sample error asymptotically with increasing
          sample size $N$, and $E_{\rm{in}}(g)$ is sufficiently small. 
          
We introduce the notions of the growth function and Vapnik-Chervonenkis 
dimension to explore the feasibility of learning using LAD. 
\begin{definition}
Let $\cH$ be a hypothesis set for the phenomenon under consideration. 
For any $h \in \cH$ and $N$ points $\xx^{(1)}, \ldots, \xx^{(N)} \in \cX$,  the $N$-tuple 
$(h(\xx^{(1)}), \ldots, h(\xx^{(N)}))$ is said to be a dichotomy. 
\end{definition}
The set of dichotomies generated by $\cH$ on the points $\xx^{(1)}, \ldots, \xx^{(N)} \in \cX$ 
is $ \cH(\xx^{(1)}, \ldots, \xx^{(N)}) 
    = \{ (h(\xx^{(1)}), \ldots, h(\xx^{(N)})) : h \in \cH \}$.  
If $\cH$ is capable of generating all possible dichotomies on 
$\xx^{(1)}, \ldots, \xx^{(N)}$, i.e., 
$\cH(\xx^{(1)}, \ldots, \xx^{(N)}) = \cB^N$, we say that 
$\cH$ {\it shatters} the set $\{ \xx^{(1)}, \ldots, \xx^{(N)} \}$. 
\begin{definition}\label{gf-def}
The growth function for a hypothesis set $\cH$ is 
\begin{equation}
m_{\cH}(N) 
= \max \left\{ \abs{\cH(\xx^{(1)}, \ldots, \xx^{(N)})} 
 : \xx^{(1)}, \ldots, \xx^{(N)} \in \cB^n \right\}.   
\end{equation}
\end{definition}
The growth function $m_{\cH}(N) \leq 2^N$
since for any $\cH$ and $\xx^{(1)}, \ldots, \xx^{(N)} \in \cB^n$, the 
set $\cH(\xx^{(1)}, \ldots, \xx^{(N)}) \subseteq \cB^N$. 
The Vapnik-Chervonenkis dimension, i.e., the VC dimension, of 
a hypothesis set $\cH$ is defined as follows. 
\begin{definition}\label{VC-dimension}
The Vapnik-Chervonenkis dimension of a hypothesis set $\cH$, denoted by 
$d_{\emph{vc}}(\cH)$, or $d_{\emph{vc}}$, is the largest value of $N$ for which 
$m_{\cH}(N) = 2^N$. If $m_{\cH}(N) = 2^N$ for all $N$, then $d_{\emph{vc}} = \infty$. 
\end{definition}
The following inequality provides an upper bound for the growth function
as a function of the VC dimension and the sample size. 
\begin{equation}\label{growth-function}
    m_{\cH}(N) \leq \sum_{i=0}^{d_{\rm{vc}}} {\binom{N}{i}}
\end{equation}
Finally, we state the VC generalization bound. 
\begin{theorem}[Theorem 2.5, page 53, \cite{mostafa}]
For any tolerance $\delta > 0$, 
\begin{equation}\label{gen-bound}
    E_{\rm{out}}(g) \leq E_{\rm{in}}(g) + \sqrt{\frac{8}{N}\ln \frac{4 m_{\cH}(2N)}{\delta}} 
\end{equation}
with probability $\geq 1 - \delta$. 
\end{theorem}

\section{LAD as a PAC learning model}
\label{LAD-Learning}
Suppose the data points in our training set $\cD$ involve $n$ 
binary features for some positive integer $n$. We use Boolean 
functions defined on $\cB^n$ to learn from such a training set. 
First, we consider the hypothesis set $\cH_n$ consisting 
of all cubic monomials in $n$ binary variables. That is 
\begin{equation}
\label{cubichypo}
\cH_n = \{x_i^{\alpha_i}x_j^{\alpha_j}x_k^{\alpha_k} 
: \alpha_i, \alpha_j, \alpha_k \in \{0,1\}, i<j<k, \mbox{ for all } i, j, k \in [n]\}. 
\end{equation}
The following theorem estimates the VC dimension of $\cH_n$.  
\begin{theorem}     
\label{cubicVC}
Let $\cH_n$ be the hypothesis set consisting of cubic monomials. Then
the VC dimension    
\begin{equation}
\label{eqVC}
d_{{\rm vc}}(\cH_n) = \Theta(\log_2 n).  
\end{equation}
\end{theorem}
\bproof Suppose $\cS \subset \cB^n$ contains $N$ vectors denoted by
\begin{equation*}
\begin{split}
\bb^{(1)} & = (b^{(1)}_1, b^{(1)}_2, b^{(1)}_3, \ldots, b^{(1)}_n)\\
\bb^{(2)} & = (b^{(2)}_1, b^{(2)}_2, b^{(2)}_3, \ldots, b^{(2)}_n)\\
\ldots    & \hspace{1.5cm} \ldots \hspace{1.5cm} \ldots\\
\bb^{(N)} & = (b^{(N)}_1, b^{(N)}_2, b^{(N)}_3,\ldots, b^{(N)}_n)
\end{split}
\end{equation*}
We set $b^{(i)}_1 = b^{(i)}_2 = 1$ for all $i \in [N]$. 
The vector corresponding to the binary representation of the non-negative 
integer $m$, where $0 \leq m \leq 2^N-1$, is denoted by 
$\yy^{(m)} = (y^{(m)}_1, \ldots, y^{(m)}_N)$. 
Our aim is to construct $\cS$ such that there exist $2^N$ cubic monomials 
in $\cH_n$ each generating a distinct vector in $\cB^N$ as the 
restriction of its truth table on $\cS$.  

The vectors 
$\yy^{(0)}$ and $\yy^{(2^N-1)}$ are generated by the monomials 
$x_1x_2x_3$ and $x_1x_2\overline{x}_3$, 
if we set $b^{(i)}_3 = y^{(0)}_i$, for all $i \in [N]$. We note that
$y^{(0)}_i = 0$, for all $i \in [N]$. 
For each non-negative integer $m$ where 
$0 \leq m \leq 2^N-1$, let $\overline{m}$ be the integer 
in the same interval that satisfies the condition  
$\yy^{\overline{m}} = \overline{\yy}^{(m)}$. 
If we set $b^{(i)}_m = y^{(m)}_i$, for all $i \in [N]$, the restrictions of the 
monomials $x_1x_2x_m$ and $x_1x_2 \overline{x}_m$ of the set $\cS$ are 
$\yy^{(m)}$ and $\overline{\yy}^{(m)} = \yy^{(\overline{m})}$, respectively.  
Therefore, if $n = 2 + 2^{N-1}$, the hypothesis set $\cH_n$ shatters a sample of 
size $N$. This means that if $n=2+2^{N-1}$, the VC dimension of $d_{{\rm vc}}(\cH_n)$
satisfies $n = 2 + 2^{N-1} \leq 2 + 2^{d_{{\rm vc}}(\cH_n) - 1}.$
Taking logarithm on both sides
\begin{equation}
\label{lbound}
d_{{\rm vc}}(\cH_n) \geq \lfloor \log_2 (n-2) + 1 \rfloor.    
\end{equation}
Since the number of distinct cubic monomials is $2^3 \times {{n}\choose{3}}$ we have 
$2^{d_{{\rm vc}}(\cH_n)} \leq 2^3 \times {{n}\choose{3}}$, 
that is
\begin{equation}
\label{rbound}
\begin{split}
d_{{\rm vc}}(\cH_n) &\leq \log_2 (2^3 \times {{n}\choose{3}}) 
 = 3 + \log_2 (\frac{n(n-1)(n-2)}{3!}).  
\end{split}
\end{equation}
Combining \eqref{lbound} and \eqref{rbound} we have 
$d_{{\rm vc}}(\cH_n) = \Theta(\log_2n)$.  
\eproof

We conjecture that 
$d_{{\rm vc}}(\cH_n) = \lfloor \log_2 (n-2) + 1 \rfloor$. 
Our experimental observations in the next section support our 
conjecture. Restricting to the asymptotic analysis we obtain 
the bounds for a larger class of functions. 

\begin{theorem}     
\label{cubicVC-t}
Let $\cH^{(t)}_n$ be the hypothesis set containing exclusively the 
DNFs consisting of $t$ cubic terms in $n$ binary variables 
where $t \leq n/3$. Then 
\begin{equation}
\label{eqVC}
d_{{\rm vc}}(\cH_n^{(t)}) = \Theta(t \log_2 n). 
\end{equation}
\end{theorem}
\bproof
Let $A(n; 3) = 2^3 \times {{n}\choose{3}}$, the number of cubic monomials in 
$n$ binary variables. The number of DNFs in $\cH^{(t)}_n$ with $t$ terms is
$B(n; t, 3) = {A(n; 3) \choose t}$.
Since $A(n; 3) = 2^3 \times {{n}\choose{3}} = \Theta(n^3)$, 
\begin{equation}
\label{estimate-1}
\begin{split}
B(n; t, 3) & 
           = \frac{A(n; 3)(A(n; 3) -1) \ldots (A(n; 3) - t + 1)}{t!}
            = \Theta(n^{3t}). 
\end{split}
\end{equation}
The VC dimension $d_{{\rm vc}}(\cH^{(t)}_n)$ satisfies
$2^{d_{{\rm vc}}(\cH^{(t)}_n)} \leq B(n; t, 3) = \Theta(n^{3t})$. Therefore, 
\begin{equation}
    \label{VC-gen-1}
        d_{{\rm vc}}(\cH^{(t)}_n) \leq  O(t\log_2n). 
\end{equation}
Since $t \leq n/3$, there are $t$ mutually exclusive subsets of binary variables
each of size three. The lower bound \eqref{lbound} obtained in Theorem~\ref{cubicVC}
implies 
\begin{equation}
    \label{VC-gen-2}
    d_{{\rm vc}}(\cH^{(t)}_n) =  \Omega(t\log_2n).
\end{equation}
Combining \eqref{VC-gen-1} and \eqref{VC-gen-2} we have 
$d_{{\rm vc}}(\cH^{(t)}_n) =  \Theta(t\log_2n)$.
\eproof

The significance of Theorem~\ref{cubicVC-t} is that if we have a data set 
with $n$ features, we are assured that the VC dimension  
$d_{{\rm vc}}(\cH_n^{(t)}) = \Theta(t\log_2 n)$. Therefore, we can start 
learning from this data set using samples of size $\Theta(t \log_2(n))$.  
Furthermore, the upper bound given in \eqref{growth-function} implies 
that if the VC dimension is finite then the growth function 
$m_{\cH}(N) = O(N^{d_{\rm{vc}}})$. Therefore, by \eqref{gen-bound}
\begin{equation}
E_{\rm{out}}(g) - E_{\rm{in}}(g) \leq 
\sqrt{\frac{8}{N}\ln \frac{4 m_{\cH}(2N)}{\delta}} 
\leq \sqrt{\frac{8}{N}\ln \frac{k (2N)^{d_{\rm{vc}}} }{\delta}}
\end{equation}
for some positive constant $k$. This implies that 
for a hypothesis class with a finite VC dimension, the in-sample error 
is an accurate approximation of the out-of sample error for large enough 
training size, $N$. As mentioned in \cite[page 56]{mostafa}, choosing 
$N = 10 \times d_{\rm{vc}}$ yields a good enough generalization to the 
out-of-sample error from the in-sample error.

\section{Experimental results}

Since LAD attempts to approximate a pdBF, we are considering the 
approximation of a random Boolean function using cubic Boolean monomials. 
In particular, we are considering the approximation of a Boolean 
function $f: \cB^{10} \rightarrow \cB$ using the hypothesis 
class $\cH_{10}$ as defined in \eqref{cubichypo}. 

We conducted an experiment wherein we chose $100$ random Boolean functions. 
For each function $f$, $50$ training sets were sampled as training data sets 
from the truth table of the Boolean function where each training set was of 
size $N$. Hypotheses in $\cH_{10}$ that corresponded to the lowest value of 
$E_{\rm{in}}$ for each training sample were considered as suitable candidates 
for approximating $f$. The corresponding $E_{\rm{out}}$ was calculated from 
the entire truth table. The algorithm of the experiment is as follows:
{
\center
\begin{algorithm}[H]
\begin{algorithmic}[1]
\STATEx{Input: Size of the training set, $N$.}
\STATE{Generate a random Boolean function $f: \cB^{10} \rightarrow \cB$ as truth table}
\STATE{Sample $f$ uniformly at random to collect $N$ samples}
\STATE{Calculate the in-sample error on N samples according to \autoref{in-error} for all functions in $\cH_{10}$}
\STATE{Identify the hypothesis function $g$ with lowest $E_{\rm{in}}(g)$}.
\STATE{Calculate $E_{\rm{out}}(g)$, from the truth tables of $f$ and $g$.}
\STATE{Store the values of in-sample and out-of-sample errors.}
\STATE{Go to Step~$2$: repeat $50$ times}
\STATE{Go to Step~$1$: repeat $100$ times}
\STATE{Plot a histogram to observe the variation in $E_{\rm{out}}(g) - E_{\rm{in}}(g)$}
\end{algorithmic}
\caption{Algorithm for the experiment.}
\label{algo-exp}
\end{algorithm}
}

If in Step~$4$ of the above algorithm, there are multiple functions having minimum $E_{\rm{in}}$, 
then all of them are considered for the following step. This was observed to be the case in 
almost all instances. 

The experiment described in Algorithm~\autoref{algo-exp} was initially 
repeated for values around $N =4$. The reason for this choice is because 
our conjectured VC dimension 
of $\cH_{10}$ is given by 
$\lfloor \log_2(10 - 2)  + 1\rfloor = 4$, 
and the given values 
will enable us to observe the connection between VC dimension and the extent to which learning 
is possible in the given experiment. 

The same experiment was then run for values $N = 10, 20, 40, 60$; this was done to observe 
the relation between $E_{\rm{out}}$ and $E_{\rm{in}}$ in the $10 \times d_{\rm{vc}}$ limit and 
to confirm if the in-sample error is indeed a good approximation of the out of sample error. 

Since we are attempting to approximate randomly generated Boolean functions $f$, the 
average value of $E_{\rm{out}}$ is going to be $0.5$. This is so because a randomly 
generated Boolean function can evaluate to $0$ or $1$ with equal probability at every input value. 
Therefore, the given experiment is not going to yield good approximations of $f$. 
This is fine as we are concerned with observing the connection between the in-sample 
and out-of-sample errors as the sample size $N$ increases.

\begin{figure}[htb]
\centering
\begin{subfigure}{0.35\textwidth}
    \includegraphics[width=\textwidth]{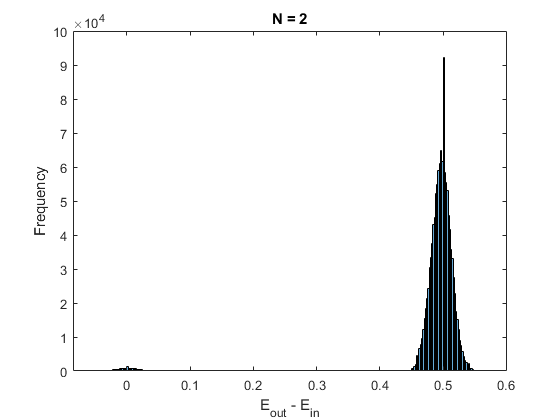}
    \caption{$N=2$}
    \label{N=2}
\end{subfigure}
\hspace{1.5cm}
\begin{subfigure}{0.35\textwidth}
    \includegraphics[width=\textwidth]{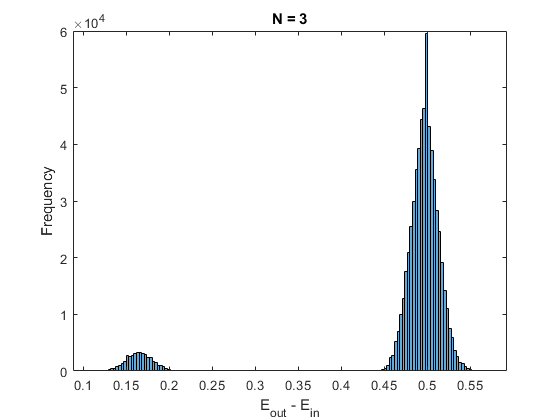}
    \caption{$N=3$}
    \label{N=3}
\end{subfigure}
\hspace{1.5cm}
\begin{subfigure}{0.35\textwidth}
    \includegraphics[width=\textwidth]{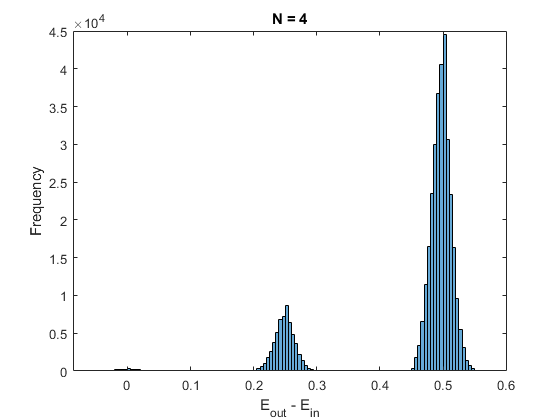}
    \caption{$N=4$}
    \label{N=4}
\end{subfigure}
\hspace{1.5cm}
\begin{subfigure}{0.35\textwidth}
    \includegraphics[width=\textwidth]{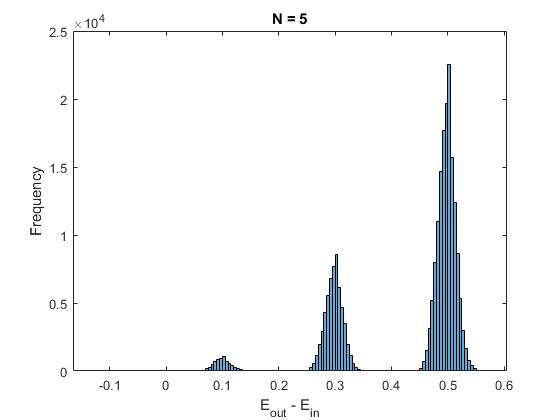}
    \caption{$N=5$}
    \label{N=5}
\end{subfigure}
        
\caption{\footnotesize{Histograms showing the distribution of $E_{\rm{out}}(g) - E_{\rm{in}}(g)$ in the neighbourhood of $d_{\rm{vc}}$.}}
\label{fig:1}
\end{figure}

The results of the initial run of the experiment can are given in \autoref{fig:1}. In the cases 
where the sample $N$ size is below $d_{\rm{vc}} = 4$, it can be seen that $E_{\rm{out}} - E_{\rm{in}}$ 
is around $0.5$ for a vast number of cases, this is due to the fact that for small sample sizes, it is possible to find a large number of hypotheses with near-zero $E_{\rm{in}}$, but many of these hypotheses will invariably be poor approximations and therefore the in-sample error is a very poor generalization for the out-of-sample error.

This situation changes as we reach $N = 4$, the (conjectured) VC dimension for this problem. There are now some situations where $E_{\rm{out}} - E_{\rm{in}} < 0.5$. In these cases, $E_{\rm{in}}$ is a relatively better generalization of $E_{\rm{out}}$. This situation improves further as one moves beyond the VC dimension in $N = 5$.

\begin{figure}[htb]
\centering
\begin{subfigure}{0.35\textwidth}
    \includegraphics[width=\textwidth]{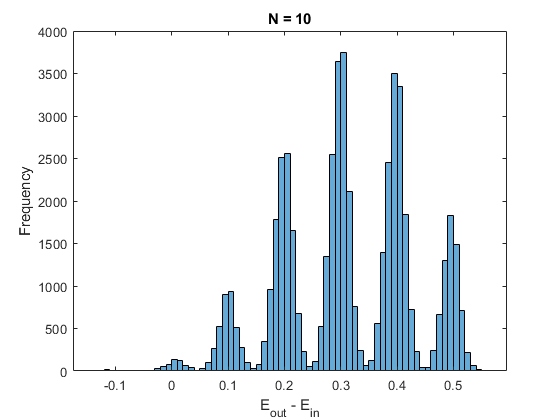}
    \caption{$N=10$}
    \label{N=10}
\end{subfigure}
\hspace{1.5cm}
\begin{subfigure}{0.35\textwidth}
    \includegraphics[width=\textwidth]{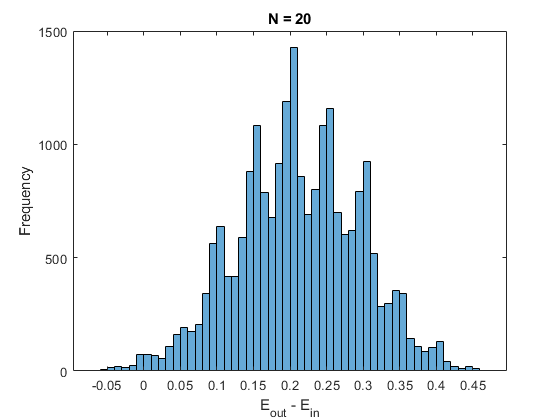}
    \caption{$N=20$}
    \label{N=20}
\end{subfigure}
\hspace{1.5cm}
\begin{subfigure}{0.35\textwidth}
    \includegraphics[width=\textwidth]{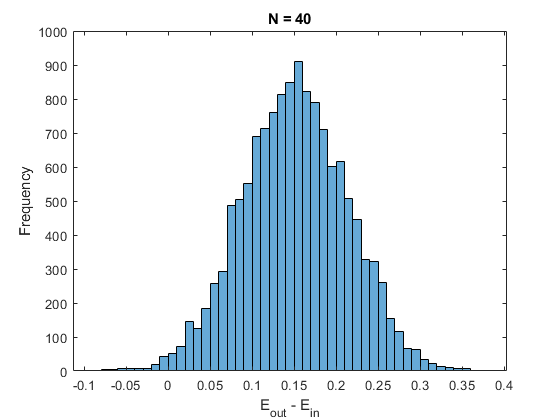}
    \caption{$N=40$}
    \label{N=40}
\end{subfigure}
\hspace{1.5cm}
\begin{subfigure}{0.35\textwidth}
    \includegraphics[width=\textwidth]{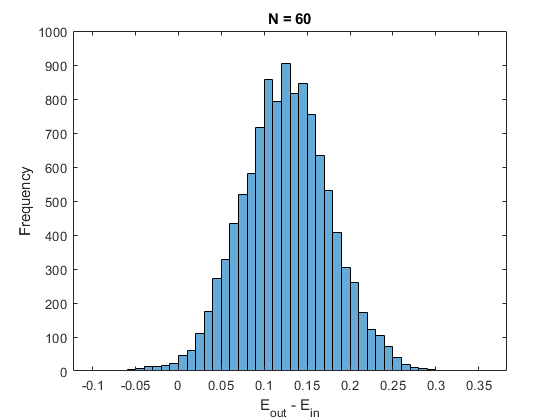}
    \caption{$N=60$}
    \label{N=60}
\end{subfigure}
        
\caption{\footnotesize{Histograms showing the distribution of $E_{\rm{out}}(g) - E_{\rm{in}}(g)$ for larger values of $N$.}}
\label{fig:2}
\end{figure}

The result of the experiment for the larger values of $N$ are given in \autoref{fig:2}, it can now be seen that lower values of $E_{\rm{out}} - E_{\rm{in}}$ are occurring with greater frequency. This enables us to establish confidence intervals for the difference between the two errors. This implies that we are in the regime of probably approximately correct (PAC) learning.

Therefore, one can state the probability for the accuracy of the estimate of the out-of-sample error with respect to $E_{\rm{in}}(g)$ for the functions belonging to $\cH_{10}$. This serves as an elementary illustration that learning becomes feasible as the size of the sample $N$, increases beyond the VC dimension.

\begin{table}[htb]
\centering
\caption{\footnotesize{Values of average in sample errors for different sample sizes}}
\label{table:1}
\scalebox{0.75}{
\begin{tabular}[c]{|c||c|c|c|c|c|c|c|c|}
\hline Sample Size ($N$) & $2$ & $3$ & $4$ & $5$ & $10$ & $20$ & $40$ & $60$ \\
\hline Avg. in-sample error ($E_{\rm{in}}$) & $0.0072$ & $0.0216$ & $0.0425$ & $0.0662$ & $0.1853$ & $0.2888$ & $0.3478$ & $0.3752$ \\
\hline
\end{tabular}}
\label{Comparison}
\end{table}

It should be noted that increasing the sample size after a point does not increase the overall accuracy of the approximation. This can be seen by reading off the values of the average in-sample error from \autoref{table:1} and observing the corresponding plot from \autoref{fig:1} or \autoref{fig:2}.

\section{Conclusion}
Logical Analysis of Data (LAD) as proposed by Peter L. Hammer demonstrates 
significantly accurate results by fitting Boolean functions to the training 
set. However, we have not found any research on incorporating LAD into the 
PAC learning framework. We initiate such an effort in this article. We 
believe that research in this direction will help us in characterizing cases 
when LAD can be used as feasible learning algorithm.  The methods presented 
here may also let us construct provably unlearnable Boolean functions.

\bibliography{lad}

\begin{thebibliography}{9}
\providecommand{\natexlab}[1]{#1}
\providecommand{\url}[1]{\texttt{#1}}
\expandafter\ifx\csname urlstyle\endcsname\relax
  \providecommand{\doi}[1]{doi: #1}\else
  \providecommand{\doi}{doi: \begingroup \urlstyle{rm}\Url}\fi

\bibitem[Hammer(1986)]{ham86}
P.L. Hammer.
\newblock Partially defined boolean functions and cause-effect relationships.
\newblock In \emph{International Conference on Multi-attribute Decision Making
  Via OR-based Expert Systems. University of Passau, Passau, Germany}, April
  1986.

\bibitem[Crama et~al.(1988)Crama, Hammer, and Ibaraki]{cra88}
Y.~Crama, P.L. Hammer, and T.~Ibaraki.
\newblock Cause-effect relationships and partially defined boolean functions.
\newblock \emph{Ann. Oper. Res.}, 16\penalty0 (1-4):\penalty0 299--325, January
  1988.
\newblock ISSN 0254-5330.

\bibitem[Alexe et~al.(2007)Alexe, Alexe, Bonates, and Kogan]{ale07}
Gabriela Alexe, Sorin Alexe, Tib{\'e}rius~O. Bonates, and Alexander Kogan.
\newblock Logical analysis of data -- the vision of {P}eter {L}. {H}ammer.
\newblock \emph{Annals of Mathematics and Artificial Intelligence}, 49\penalty0
  (1):\penalty0 265--312, Apr 2007.
\newblock \doi{10.1007/s10472-007-9065-2}.
\newblock URL \url{https://doi.org/10.1007/s10472-007-9065-2}.

\bibitem[Chauhan et~al.(2022)Chauhan, Mahmoud, Gangopadhyay, and
  Gangopadhyay]{ChauhanMGG22}
Sneha Chauhan, Loreen Mahmoud, Sugata Gangopadhyay, and Aditi~Kar Gangopadhyay.
\newblock A comparative study of lad, {CNN} and {DNN} for detecting intrusions.
\newblock In Hujun Yin, David Camacho, and Peter Ti{\~{n}}o, editors,
  \emph{Intelligent Data Engineering and Automated Learning - {IDEAL} 2022 -
  23rd International Conference, {IDEAL} 2022, Manchester, UK, November 24-26,
  2022, Proceedings}, volume 13756 of \emph{Lecture Notes in Computer Science},
  pages 443--455. Springer, 2022.
\newblock \doi{10.1007/978-3-031-21753-1\_43}.
\newblock URL \url{https://doi.org/10.1007/978-3-031-21753-1\_43}.

\bibitem[Crama and Hammer(2011)]{CramaH}
Yves Crama and Peter~L. Hammer.
\newblock \emph{Boolean Functions - Theory, Algorithms, and Applications},
  volume 142 of \emph{Encyclopedia of mathematics and its applications}.
\newblock Cambridge University Press, 2011.
\newblock ISBN 978-0-521-84751-3.
\newblock URL
  \url{http://www.cambridge.org/gb/knowledge/isbn/item6222210/?site\_locale=en\_GB}.

\bibitem[Boros et~al.(2011)Boros, Crama, Hammer, Ibaraki, Kogan, and
  Makino]{DBLP:journals/anor/BorosCHIKM11}
Endre Boros, Yves Crama, Peter~L. Hammer, Toshihide Ibaraki, Alexander Kogan,
  and Kazuhisa Makino.
\newblock Logical analysis of data: classification with justification.
\newblock \emph{Ann. Oper. Res.}, 188\penalty0 (1):\penalty0 33--61, 2011.
\newblock \doi{10.1007/s10479-011-0916-1}.
\newblock URL \url{https://doi.org/10.1007/s10479-011-0916-1}.

\bibitem[Valiant(1984{\natexlab{a}})]{Valiant84}
Leslie~G. Valiant.
\newblock A theory of the learnable.
\newblock \emph{Commun. {ACM}}, 27\penalty0 (11):\penalty0 1134--1142,
  1984{\natexlab{a}}.
\newblock \doi{10.1145/1968.1972}.
\newblock URL \url{https://doi.org/10.1145/1968.1972}.

\bibitem[Valiant(1984{\natexlab{b}})]{Valiant-stoc-84}
Leslie~G. Valiant.
\newblock A theory of the learnable.
\newblock In Richard~A. DeMillo, editor, \emph{Proceedings of the 16th Annual
  {ACM} Symposium on Theory of Computing, April 30 - May 2, 1984, Washington,
  DC, {USA}}, pages 436--445. {ACM}, 1984{\natexlab{b}}.
\newblock \doi{10.1145/800057.808710}.
\newblock URL \url{https://doi.org/10.1145/800057.808710}.

\bibitem[Abu-Mostafa et~al.(2012)Abu-Mostafa, Magdon-Ismail, and Lin]{mostafa}
Yaser~S. Abu-Mostafa, Malik Magdon-Ismail, and Hsuan-Tien Lin.
\newblock \emph{Learning From Data}.
\newblock AMLBook, 2012.
\newblock ISBN 1600490069.

\end{thebibliography}
\end{document}